\documentclass{ieeeaccess}
\usepackage{cite}
\usepackage{amsmath,amssymb,amsfonts}
\usepackage{algorithmic}
\usepackage{graphicx}
\usepackage{textcomp}
\usepackage{lipsum}
\usepackage{systeme,mathtools}
\usepackage{float}
\usepackage{placeins}
\usepackage{flushend}
\usepackage{booktabs}
\usepackage{color,soul}
\usepackage{colortbl}
\usepackage{hyperref}
\usepackage{url}

\definecolor{yellow}{rgb}{1,1,0}
\newcolumntype{a}{>{\columncolor{yellow}}c}
\newcolumntype{b}{>{\columncolor{yellow}}p{2.5cm}}

\definecolor{cerulean}{rgb}{0.0, 0.48, 0.65}
\definecolor{carnelian}{rgb}{0.7, 0.11, 0.11}
\definecolor{buff}{rgb}{0.94, 0.86, 0.51}
\definecolor{amethyst}{rgb}{0.6, 0.4, 0.8}
\definecolor{asparagus}{rgb}{0.53, 0.66, 0.42}

\def\BibTeX{{\rm B\kern-.05em{\sc i\kern-.025em b}\kern-.08em
    T\kern-.1667em\lower.7ex\hbox{E}\kern-.125emX}}
\begin{document}
\history{Date of publication xxxx 00, 0000, date of current version xxxx 00, 0000.}
\doi{10.1109/ACCESS.2017.DOI}

\title{Graph learning in robotics: a survey}
\author{\uppercase{Francesca Pistilli}\authorrefmark{1}, \IEEEmembership{Student Member, IEEE} and
\uppercase{Giuseppe Averta\authorrefmark{1},
\IEEEmembership{Member, IEEE}}
\address[1]{Department of Control and Computer Engineer, Polytechnic of Turin, Turin, Italy (e-mail: name.surname@polito.it)}
}
\markboth
{Pistilli \headeretal: Graph learning in robotics: a survey}
{Pistilli \headeretal: Graph learning in robotics: a survey}

\corresp{Corresponding author: Francesca Pistilli (e-mail: name.surname@polito.it).}

\begin{abstract}
Deep neural networks for graphs have emerged as a powerful tool for learning on complex non-euclidean data, which is becoming increasingly common for a variety of different applications. Yet, although their potential has been widely recognised in the machine learning community, graph learning is largely unexplored for downstream tasks such as robotics applications.
To fully unlock their potential, hence, we propose a review of graph neural architectures from a robotics perspective. 
The paper covers the fundamentals of graph-based models, including their architecture, training procedures, and applications. It also discusses recent advancements and challenges that arise in applied settings, related for example to the integration of perception, decision-making, and control. Finally, the paper provides an extensive review of various robotic applications that benefit from learning on graph structures, such as bodies and contacts modelling, robotic manipulation, action recognition, fleet motion planning, and many more. This survey aims to provide readers with a thorough understanding of the capabilities and limitations of graph neural architectures in robotics, and to highlight potential avenues for future research.
\end{abstract}


\titlepgskip=-15pt

\maketitle

\section{Introduction}
\label{sec:introduction}

Over the last few years, we observed an exponential growth of deep learning methods for a variety of different data modalities and applications. Thanks to its closeness to real world applications, robotics represents the perfect ultimate harbour where deep learning methods find their downstream task. 

\begin{figure*}[t]
    \centering
    \includegraphics[width=0.75\textwidth]{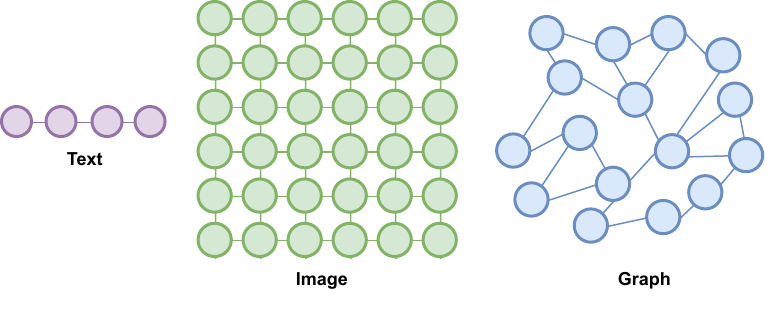}
    \caption{Schematics of the information representation encoded in the form of images, text and graphs. Although being by far the most common data structures to train artificial neural architectures, images and ordered series of tokens cannot handle unstructured data distribution. To address this challenge, graph structures constitute the primary mathematical representation for an efficient and effective data representation.}
    \label{fig:img_txt_net}
\end{figure*}

Notable examples are the field of robot vision, and the upcoming trend of foundational multi-modal models for task and motion planning, where images are coupled with other information sources such as text and audio. However, in many practical cases, a mere representation of knowledge through bi-dimensional pixels or temporal sequence of tokens may not be sufficient to properly convey the information. Three-dimensional visual data, functional and geometrical relationships, and interaction between multiple agents and objects - for example - require more complex and unstructured data representation (see Fig. \ref{fig:img_txt_net}). One popular approach to handle these cases is through graphs encoding, where information is represented as an organised set of nodes and edges, embedding relevant features of the task. 

Noteworthy, very often we observe a significant delay between when new methods are proposed in the machine learning and computer vision communities, and their application in robotics-oriented use-cases. Therefore, the purpose of this survey is to lay down the current state of the art on graph learning, with specific focus on their potential application on robotic tasks, with the aim of collecting the available knowledge at the moment, and showcase the potential of graph theory and learning, to foster future developments of intelligent machines able to learn on complex data structures.   

\subsection{Learning from unstructured data} 

Graph neural networks are powerful mathematical tools for knowledge representation, able to encode structured and unstructured information in a convenient fashion, through nodes which model entities (objects, functional elements, robots), and edges that encode their spatial, temporal or functional relationship. Graphs are commonly used to model a variety of different elements, such as molecules \cite{kearnes2016molecular} for drug discovery, physics simulations \cite{sanchez2020learning}, scientific citations \cite{scarselli2008graph} or social networks \cite{fan2019graph}. 
In computer vision, this architecture has emerged as a promising tool to deal with data that lie on irregular domains, such as point clouds \cite{simonovsky2017dynamic, wang2019dynamic}, to uncover non-local similarities in the data \cite{valsesia2020deep}, and video understanding\cite{sigurdsson2016hollywood, nagarajan2020ego}. 

Thanks to these promising results, Graph Neural Networks (GNNs) have gained the attention of the robotic community, being proposed as key technological solutions for more and more new tasks and applications. As noticeable examples, it is worth mentioning grasping and manipulation, in which robots are usually tasked to process as input three-dimensional data, for which graphs can be used to provide an efficient yet effective modelling tool. Another relevant scenario is the use of graph architectures to capture and model spatial and functional keypoints of environments or work-spaces, allowing robots to better understand the environment wherein, how to interact safely with objects and zones, and how to plan meaningful tasks.  

GNNs also lend themselves well to reinforcement learning, as they can learn to predict future states based on current state and action signals. In robotics, this brings the benefit that GNNs can effectively plan robot actions sequences taking into account current and future states to select the best succession of action. In addition, GNNs can help robots learn how to move and behave in more complex environments by providing more detailed models of their surroundings. Deep Reinforcement Learning (DRL) has been a resourceful tool for improving decision-making tasks and control problems in robot automation \cite{chen2020autonomous,chen2021zero,luo2019multi}. However, in many cases DRL suffers of limited generalisation capability, with limits a seamlessly adaptation to new scenarios, different than the ones experienced at training time \cite{packer2018assessing}. The ability to generalise to novel scenarios it is of paramount importance for many practical applications, and to shorten the distance between sim-to-real simulations. To address this limitation, several projects demonstrated significant benefits in the use of GNNs for learning robust policies across scenarios \cite{chen2020autonomous,chen2021zero,luo2019multi}, benefiting from the intrinsic generalisation ability of graphs encoding. 

\subsection{Survey Methodology}
To ensure an high quality selection of papers for our survey, we first employed Elsevier Scopus \footnote{\url{https://www.scopus.com}} as search engine, using as keywords deep learning, robotics, graph learning and graph neural networks. This search yielded a first list of documents, from which we then selected only the manuscript published in journals with ranking Q2 or above on ScimagoJR \footnote{\url{https://www.scimagojr.com}} for Computer Science or Control and Systems Engineering and indexed in Web of Science. For conferences papers, instead, we referred to the CORE \footnote{\url{http://portal.core.edu.au/conf-ranks/}} conferences ranking system, which provides an assessment of the major conferences in the field, excluding papers published in venues ranked B or below (i.e. we keep only A* and A -- the top 22\%). We then refined the bibliographic search by manually checking for other relevant papers exploiting suggested similar publication in Google Scholar (still matching the above constraints). This process yielded the selection revised in this survey.

\subsection{How to navigate this survey}
This paper is organised as follows. In Section II, we provide a concise review of the theoretical background behind graph learning, with specific focus on different graph convolution formulations. Following, we explore the use of GNNs in robotics applications. First, in Section III we showcase the application of GNNs in the context of single-agent settings. With such definition we refer to cases where graphs are used to model one single item, which encompasses three main cases: modelling of objects (Section IIIa), modelling of hands-objects interaction (section IIIb) and modelling of human behaviour (Section IIIc). Then, Section IV discusses multi-agent settings, where graphs are used to model the interaction between multiple items. The section is further divided in Task and motion planning (Section IVa) and multi-robot exploration and navigation (Section IVb). We also provide a section where we discuss strengths and limitations of current approaches including GNNs in their pipeline, and finally a conclusion with a forecast of future perspectives. 
All the parts are self-contained, allowing the curious reader to directly refer to a specific section of interest. 

\section{Preliminaries on Graph Neural Networks}
In the last few years, learning on graph structures has gained significant relevance in the community, thanks to their intrinsic capability to easily process data lying on irregular domains, such as three-dimensional point clouds, spatio-temporal and functional relationships \cite{bacciu2020gentle, wu2020comprehensive,yuan2022explainability,zhou2020graph}.
Similarly to what Convolutional Neural Networks (CNNs) did for standard images, the community has devoted significant effort in defining convolution-based aggregation mechanisms for graph structures.

\begin{figure}[t]
    \centering
    \includegraphics[width = 0.37\columnwidth]{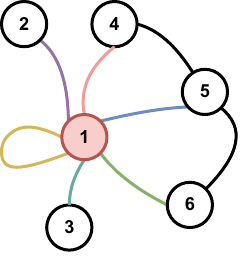}
    \quad \quad
    \includegraphics[width = 0.37\columnwidth]{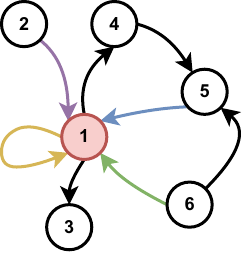}
    \caption{Undirected (left) and a directed (right) graph representation. The neighbourhood of node 1 in highlighted through coloured edges in both graphs. For the undirected graph, the neighbourhood is composed by the nodes that are connected to node 1 : node 2, node 3, node 4, node 5, node 6. Instead, for the directed one, the neighbourhood is composed only by the nodes that have incoming edges to node 1: node 2, node 5, node 6. It is worth mentioning that in some cases graphs may also have edges starting and ending on the same node (self-loop), which may be relevant for some graph convolution formalisation. }
    \label{fig:graphs}
\end{figure}

A graph convolution operation can be formulated over two different domains, spectral or spatial. The first family of methods \cite{bruna2013spectral, henaff2015deep,defferrard2016convolutional,kipf2016semi, monti2018motifnet} usually exploits graph Fourier transform, eventually complemented with polynomial approximations to reduce the computational burden \cite{defferrard2016convolutional, kipf2016semi}. 
Among these, it is worth mentioning the Graph Convolutional Network (GCN) \cite{kipf2016semi}, which demonstrated notable results for semi-supervised problems, such as semi-supervised node or multi-class classification. However, a critical limitation of this formulation is the inability to generalise the learned filters, computed over the spectrum of the graph Laplacian, to a variable graph structure.
\begin{figure*}[t]
    \centering
    \includegraphics[width=0.9\textwidth]{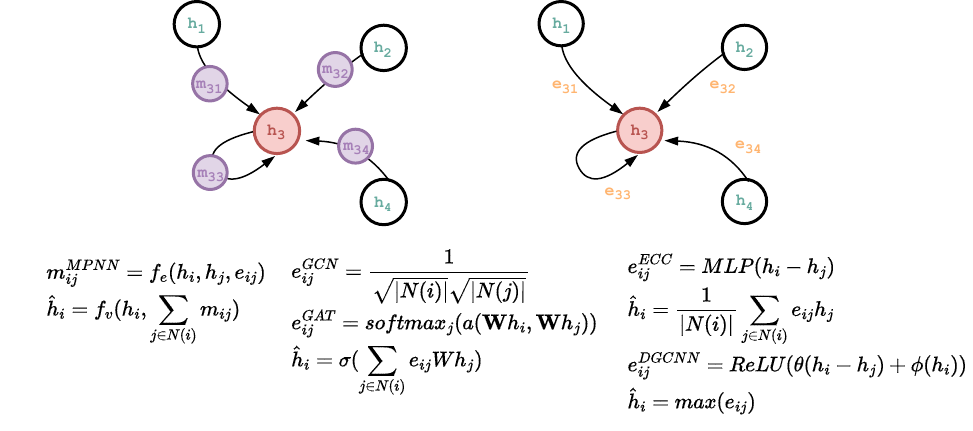}
    \caption{Graph convolutional operations with different information sharing mechanism. MPNN \cite{gilmer2017neural} implements message passing conditioned by edge attributes. GCN \cite{kipf2016semi} is reported in its equivalent spatial formulation where it implements a simple aggregation of the neighbouring nodes. With a similar operation,  GAT \cite{velivckovic2017graph} exploits self-attention to weight the contributions of the nearest neighbours. Instead ECC \cite{simonovsky2017dynamic} implements an edge-depended weighting function to weight the contribution of each node in the neighbourhood. DGCNN \cite{wang2019dynamic} aggregate the information of the neighbourhood by means of learned edge features.}
    \label{fig:gnns_formalization}
\end{figure*}

The second class of approaches defines the graph-convolution operation in the spatial domain. In this scenario, the graph convolution is defined as a local, i.e. computed over a neighbourhood, weighted aggregation of signals. Since it is defined at the neighbourhood level, this formulation is suitable for any type of signal that can be defined over a graph, even with a variable graph structure. Several definitions are present in the literature \cite{monti2017geometric, velivckovic2017graph, hamilton2017inductive, simonovsky2017dynamic, verma2018feastnet, xu2018powerful, wang2019dynamic, wang2019graph, corso2020principal}. These often differ on the computation of the weights used in the aggregation. For example, many formulations  use scalar weights \cite{monti2017geometric,velivckovic2017graph, verma2018feastnet} or matrices \cite{hamilton2017inductive,xu2018powerful, wang2019dynamic,corso2020principal} independent from the input data. 
On the other side, \cite{simonovsky2017dynamic} proposes edge-dependent matrices as weights, providing an operation with more representational power. For a compact overview, the reader could refer to Fig. \ref{fig:gnns_formalization} for different graph convolution formulations and to Tab. \ref{tab:comparison} for a comparison between implementations.

\subsection{Graph Signal theory backgrounds} \label{sec:gnn_back}

Given a graph $\textit{G} = \{\textit{V}, \textit{E}\}$, we refer to $\textit{V}$ as the set of nodes, with cardinality $M = |\textit{V}|$, and $\textit{E}$ as the set of edges. 
Two nodes, $n_i$ and $n_j$ are said to be connected only if there exists an edge $e_{i, j} \in \textit{E}$ between them. The graph is also undirected if $e_{i, j} = e_{j, i}, \quad \forall e_{i,j} \in \textit{E}$, otherwise the order of the indices reflects the direction of the graph and in general $e_{i, j} \neq e_{j, i}$. Each node or edge can be associated with a feature vector, $\textbf{F}_v \in \mathbb{R}^{d_v}$ or $\textbf{F}_e \in \mathbb{R}^{d_e}$. The neighbourhood of the $i^{th}$ node is defined as $N(i) = \{j | e_{i, j} \in \textit{E}\}$, and corresponds to the set of nodes directly connected to the $i^{th}$ node. In Fig. \ref{fig:graphs} it is reported an example of an undirected and a directed graphs, where the neighbourhood of a node is highlighted.

The connectivity of the graph can be represented through the adjacency matrix $\textbf{A} \in \mathbb{R}^{M,M}$, where element $A_{i,j} = 1$ if exists the corresponding edge $e_{i,j} \in \textit{E}$, and zero otherwise. Note that, if the graph is undirected, the corresponding adjacency matrix is symmetric. From the connectivity matrix it is possible to define the diagonal degree matrix $\textbf{D}$, where the diagonal element $d_{i,i}$ is equal to the sum of all the edges, incoming and outgoing if in presence of a directed graph, relative to the $i^{th}$ node. In some cases, edges could be weighted by specific values $w_{i,j}$, and the corresponding element of the adjacency matrix becomes $A_{i,j} = w_{i,j} $ if $e_{i,j} \in \textit{E}$, and zero otherwise.

Finally, it is worth recalling the definition - for undirected graphs - of the Laplacian matrix $\hat{\textbf{L}} = \textbf{D} - \textbf{A}$, and of its normalised formulation $\textbf{L} = \textbf{I} - \textbf{D}^{-\frac{1}{2}} \textbf{A} \textbf{D}^{-\frac{1}{2}}$. Since the normalised graph Laplacian is a semi-define positive matrix, it can be decomposed as $\textbf{L} = \textbf{U}\textbf{\Lambda} \textbf{U}^T$. Given a generic graph signal $f: \textit{V} \rightarrow \mathbb{R}$, which can be represented in vector form as $\textbf{x} \in \mathbb{R}^M$ where each element $\textbf{x}_i$ is the function evaluated in the $i^{th}$ node, it is possible to define the graph Fourier transform $\mathcal{F}$ of the signal $\textbf{x}$ and its inverse $\mathcal{F}^{-1}$ as: 
\begin{align}  
    & \mathcal{F}(\textbf{x}) = \hat{\textbf{x}} = \textbf{U}^T\textbf{x} = \sum_{i=0}^{M-1} \textbf{x}_i u_{i} ,\\
    & \mathcal{F}^{-1}(\hat{\textbf{x}}) = \textbf{x} = \textbf{U} \textbf{x} = \sum_{i=0}^{M-1} \hat{\textbf{x}}_i u_{i}.
\end{align}

The graph Fourier transform projects $\textbf{x}$ into an orthonormal space where the eigenvectors of the graph Laplacian matrix are the basis of the new space. Thanks to their definition, the graph Laplacian operators family captures the differences between a signal at a node and its surroundings, and specifically the graph Laplacian is often used as a measure of signal smoothness. Using the graph Fourier transform, it is possible to filter the signal either in the spectral or in the spatial domain, and define a graph convolutional operation. 

\subsubsection{Spectral Graph Convolutions}
In the spectral domain, the filtering operation corresponds to a multiplication between the graph signal $\textbf{x}\in \mathbb{R}^M$ and the filter $\textbf{g} \in \mathbb{R}^M$:
\begin{equation}
    \textbf{x} *_G \textbf{g} = \mathcal{F}^{-1} (\mathcal{F}(\textbf{x}) \odot \mathcal{F}(\textbf{g})) =  \textbf{U}( (\textbf{U}^T\textbf{x}) \odot (\textbf{U}^T\textbf{g}) ), 
\end{equation}
where $\odot$ is the Hadamard (element-wise) product. 

Considering a filter in the form $\textbf{g}_{\theta} = diag(\textbf{U}^T\textbf{g})$, the spectral graph convolution can be simplified to:
\begin{equation}
    \textbf{x} *_G \textbf{g} = \textbf{U}\textbf{g}_{\theta}\textbf{U}^T\textbf{x}.
\end{equation}
The distinct definitions of spectral graph convolution differs for the choice of the filter $\textbf{g}_{\theta}$. 
Bruna et al. \cite{bruna2013spectral} proposes a first repurpose of classic CNNs to the graph domain, and introduces a spectral convolution where the graph convolutional filter is modelled as a learnable diagonal matrix $\textbf{g}_{\theta} = \Theta$ obtaining a simple yet effective formulation.
Defferrard et al. \cite{defferrard2016convolutional}, instead, proposes an efficient formulation approximating the filters as  Chebyshev polynomials $\textbf{g}_{\theta} = \sum_{k=0}^{K-1} \theta_k T_k(\Tilde{L})$. $T_k(\Tilde{L})$ is the Chebychev polynomial of order $k$ evaluated at the scaled Laplacian $\Tilde{L}=\frac{2\textbf{L}}{\lambda_{max}} - \textbf{I}$ and $\theta$ is a learnable vector of length $k$. Recall that the Chebychev polynomials are recursively defined as $T_k(\textbf{x}) = 2\textbf{x}T_{k-1}(\textbf{x}) - T_{k-2}(\textbf{x})$ with $T_0(\textbf{x}) = 1$ and $T_1(\textbf{x}) = \textbf{x}$.
Given an input signal $\textbf{x}^t$ at step $t$, the output signal $\textbf{x}^{t+1}$ after the graph convolution is:
\begin{equation}
    \textbf{x}^{t+1} = \sum_{k=0}^{K-1} \theta_k T_k(\Tilde{L}) \textbf{x}^t.
\end{equation}
An interesting observation is that the filters are localised in a spatial neighbourhood.
Such formulation is further simplified by Kipf et al. \cite{kipf2016semi}, which proposes a first-order approximation of the Chebychev polynomials and exploits the normalised graph Laplacian build upon the adjacency matrix with self-loop connections $\hat{A}$ instead of the scaled Laplacian:
\begin{equation}
    \textbf{x}^{t+1} = \alpha (I - L) \textbf{x}^t,
\end{equation}
where $\alpha$ is a learnable scalar parameter. Such formulation consists in simply taking a weighted aggregation of the signals in the neighbourhood of each points with weights proportional to the connectivity matrix. 

Such models have the advantage of providing a formulation for filters localised in the neighbourhood space, with a limited number of learnable parameters and independent from the graph size. Nevertheless, they are linked to a fixed graph structure, given by the graph Laplacian, and therefore are not able to generalise to other configurations. Furthermore, they are not robust to perturbations that would change the basis functions, and they require a computational cost of $O(N^2)$.

\begin{table*}[t]
    \begin{center}
    \begin{tabular}{ ccccc }
      Name & Type & Weights computation & Edges & Robotic Applications \\ 
     \toprule
     Bruna et al. \cite{bruna2013spectral} & Spectral & lernable diagonal matrix & w/o attributes & - \\
     
     Defferard et al. \cite{defferrard2016convolutional} & Spectral & learnable parameters & w/o attributes & \cite{deng2022graph} \cite{luo2019multi} \\
          
     GCN \cite{kipf2016semi} & Spectral & learnable parameters $\propto$ connectivity & w/o attributes & \cite{murali2020taskgrasp} \cite{nagarajan2020ego}\cite{dessalene2020egocentric} \cite{dessalene2021forecasting}\\

     GNN \cite{scarselli2008graph} & Spatial & learnble paramters & w/o attributes & \cite{pfaff2020learning} \cite{huang2023defgraspnets} \cite{wang2018nervenet} \cite{prorok2018graph} \cite{chen2019autonomous} \\

     MPNN \cite{gilmer2017neural} &  Spatial & learnable matrices & w/ attributes & -  \\
     
     GatedGCNN \cite{li2015gated} & Spatial  & lernable matrices & support multiple types & \cite{ding2022visual}  \\
     
     Monti et al. \cite{monti2017geometric} & Spatial & edge weighting function  & w/ attributes & -  \\
     
     GraphSAGE \cite{hamilton2017inductive} & Spatial & MLP & w/o attributes &  -\\

     GAT \cite{velivckovic2017graph} & Spatial &  self-attention & w/o attributes & \cite{li2021message} \cite{wang2020learning}  \\
         
     ECC \cite{simonovsky2017dynamic} & Spatial &  edge weighting MLP & w/ attributes & - \\
     
     DGCNN \cite{wang2019dynamic} & Spatial &  MLP & w/ attributes & - \\
     \bottomrule
    \end{tabular}
    \end{center}
    \caption{Comparison between different Graph Convolutional Neural Architectures. For each implementation, we report few core characteristics: the type (spectral vs. spatial), how weights are computed, the availability of attributes for edges and a list of relevant papers where these have been used.}
    \label{tab:comparison}
\end{table*}

\subsubsection{Spatial Graph Convolutions}
As discussed before, an alternative representation of the graph convolution operation may be also formulated in the spatial domain. The filtered signal $\textbf{x}_i$ at the $i^{th}$ node is computed as the linear combination of the signal itself and its k-hop local neighbourhood $N_k(i)$. The k-hop neighbourhood of node $i$ is defined as the collection of nodes that have the shortest path distance from the $i^{th}$ node less or equal to k. In this survey, we assume the standard practice of considering the 1-hop neighbourhood of a node - if not specified otherwise - and $N_1(i) = N(i)$. Therefore, it is possible to define the spatial graph filtering as:
\begin{equation}
    \textbf{x}_i^{t+1} = W_{i,i}\textbf{x}^t_i + \sum_{j\in N_k(i)} W_{i,j}\textbf{x}_j^t,
\end{equation}
where the first part refers to the contribution of the $i^{th}$ node itself, also called self-loop, and the second to the contribution of the surroundings points.
The spatial approach shares similar principle ideas with propagation and message passing of recurrent neural networks, since they both propagate and exchange information within the neighbourhood structure. One of the first message passing models is presented in \cite{scarselli2008graph}, where the signal function at node $i$ is updated at each time step $t$ with the aggregation of the information, also called messages, from each node $j$ in its neighbourhood. It implements an extension of previous recurrent graph neural networks to more general scenarios, considering different types of graphs. The proposed message passing mechanism can be formulated as follow:
\begin{equation}
    \textbf{x}_i^{t+1} = u(F_{v,i} , \sum_{j \in N(i)} m(\textbf{x}_i^t, F_{v,i}, F_{v,j}, F_{e, (i,j)})),
\end{equation}
where $m$ is the message passing function, $u$ the update function, $F_{v,i}$ the feature vector associated to the $i^{th}$ node and $F_{e, (i,j)}$ the feature vector associated to the edge that connects nodes $i$ and $j$. 

Over the years, several models based on message passing have been developed \cite{duvenaud2015convolutional, kearnes2016molecular, schutt2017quantum, gilmer2017neural}, where different message passing and update functions are proposed. 
Among the others, it is worth mentioning Gilmer et al. \cite{gilmer2017neural}, which proposes a general model called Message Passing Neural Network (MPNN), that implements a spatial graph convolution by means of message passing, where the information is shared between nodes conditioned by the edge attribute. Instead, the GatedGCN proposed in \cite{li2015gated} includes a Gated Recurrent Unit (GRU) \cite{cho2014learning} and the support for different edge types and directions.
Similarly, Monti et al. \cite{monti2017geometric} presents a Gaussian mixture model that exploits continuous edge attributes. 

A spatial graph convolutional neural network can be interpreted as a local weighted aggregation of points.
Locality reflects the computation performed only at neighbourhood level, while weights are used to scale different contributions. The aggregation function should keep the permutation invariance to the node ordering, and therefore it is usually a sum, mean or max function. In the literature, researchers proposed several spatial graph convolutional formulations which differ in one of the three main characteristics of the function: computation of the neighbourhood, weighting function and type of aggregation. GraphSAGE \cite{hamilton2017inductive} proposes to sub-sample the neighbours, aggregate their information with the centre node and then project into the feature space, promoting similar embedding in close nodes. Graph attention networks (GAT) \cite{velivckovic2017graph}, instead, exploits a self-attention mechanism to compute the weight for each neighbours. Edge-Conditioned Convolution (ECC) \cite{simonovsky2017dynamic} proposes an edge-dependent convolution definition. In particular, the neighbouring contributions are weighted by the output of a multi-layer perceptron that takes as input the corresponding edge labels, commonly defined as the difference between the features associated to each neighbours and the centre point. Instead Dynamic Graph CNN (DGCNN) \cite{wang2019dynamic} proposes a novel formulation called EdgeConv to generate edge features to describe the relationships between a centre point $i$ and its nearest neighbours. The edge features are defined as the output of a multi-layer perceptron that takes as input the features associated to each node $j \in N(i)$ and the centre node $i$ and are directly aggregated together to obtain the new feature vector of the centre node. Remarkably, DGCNN \cite{wang2019dynamic} first introduces the concept of dynamic graph construction, i.e. the edges and consequently the neighbourhood of each node is updated after several graph convolutional layers. 

As we can observe from the previous description, one important difference between graph convolution formulations is the contribution of the neighbourhood: GraphSAGE \cite{hamilton2017inductive} considers each contribution equal, GCN \cite{kipf2016semi} uses a fixed defined a-priori weight per neighbour, while ECC \cite{simonovsky2017dynamic} proposes edge dependent weighting contributions.

\section{GNNs for Single-Agent systems}
In this survey, we refer to "Single-Agent Systems" either when we review papers dealing with only one intelligent system (single-robot or human), or passive bodies (objects modelling). All the above scenarios 
share the common principle of using graphs to encode or guide the action of a single robot. For the sake of clarity, we further split the topic in: i) modelling of bodies, where graphs are used to encode the dynamic or kinematic behaviour of a passive object (usually compliant, such as in \cite{li2018learning} and \cite{pfaff2020learning}); ii) modelling of robots interacting with objects, mostly for grasping and manipulation purposes (e.g. \cite{murali2020taskgrasp}); iii) modelling of human behaviour and human-centric environments (e.g. \cite{nagarajan2020ego}). 





\subsection{Modelling of bodies}\label{sec:bodies}

When we interact with a compliant object, for instance during grasping or manipulation, very likely it will suffer a non-linear deformation which depends on the force and torques applied by fingers and the external environment, and on the dynamic behaviour of the body itself. However, modelling how objects deform when subject to external forces is an open problem, which poses several technical and theoretical challenges, such as: i) the high dimensionality of the configuration space, ii) the non-linear dynamics of deformable materials, and iii) possible self-occlusion in visual perception. 

One promising approach relies on the discretisation of the continuous body representation, through the identification of a finite number of keypoints. The spatial and kinetic relationship between keypoints can be easily represented through edges connecting neighbouring keypoints. Such graph encoding can be used to provide a compact, yet approximated, representation of the body shape and, potentially, its dynamic behaviour \cite{sanchez2020learning, li2018learning}. 
Message passing between nodes of a graph is a powerful tool able to learn useful spatio-temporal information on the interaction between elemental components of compliant bodies. 
Recently, learning control policies from deformable objects keypoints has gained attention \cite{li2018learning, ma2022learning, wang2022offline, shi2022robocraft, deng2022graph, deng2023learning} thanks to their effective low-dimensional alternative representation.

One of the first works going in this direction is proposed in \cite{li2018learning}, where the authors move from the observation that particle-based simulators \cite{macklin2014unified} are widely used to model complex dynamics, but very often imperfect because of a set of approximations which does not scale well in real-world scenarios. To tackle this problem, 
in \cite{li2018learning} the authors proposed to learn a particle-based simulator with a network able to model the dynamic interaction between particles of an object through a graph neural network. This approach comes with the relevant benefit of being suitable for a wide range of objects (rigid, soft and fluids), and demonstrated interesting capabilities in inferring an inductive bias on the type of particles, useful to quickly adapt to unknown environments.  
The proposed interaction model takes inspiration from \cite{battaglia2016interaction}, extending it to particle-based dynamics. Interestingly, the authors introduce a dynamic graph construction, where nodes are all the particles of an object, and edges - representing the interaction between particles - are dynamically updated in time to grant a trade-off between efficiency and effectiveness.
This is an interesting design choice which - compared to the alternative fixed fully connected graph - has the advantages of being more efficient and more effective. In real-world physical systems, since not all the points would interact with all the others, it is reasonable to assume a distance-based neighbourhood, and they involve discontinuous functions that can not be taken into account with a fixed graph construction.
They also introduce a message-passing mechanism \cite{li2019propagation} which is object-type dependent, meaning that different rules are applied for rigid bodies, deformable objects, and fluids. The same dependencies are applied to the graph construction, leading to different graphs for different types of objects.

Although particle-based approaches demonstrated interesting results in terms of accuracy, in many cases they also require a large number of particles to properly approximate a realistic object, which may result in excessive computational cost, prohibitive for real-time scenarios. Alternative solutions, which mitigate this issue, usually approximate the overall complexity by focusing only on a strategic subset of points, also known as keypoints, easy to detect and sufficient to estimate the object dynamics. As an example, G-DOOM \cite{ma2022learning} extracts keypoints from depth images via unsupervised learning and uses such data as nodes of a graph. A graph convolutional neural network is then used to capture the underlying geometry and abstract interactions between keypoints. They also introduce a recurrent structure in the architecture to track keypoints over time, with the aim of handling potential keypoints occlusions. 

The findings discussed above suggest that keypoints-based discretisation seems to be the right choice to enable a low-latency simulation of soft bodies. However, being an approximation of the real world, it is worth remarking that this will also introduce a non-negligible discrepancy between simulated and real-world data distribution (sim-to-real gap) which, in some cases, may produce important deviations from the simulated environment. To address this issue, \cite{wang2022offline} applies GNNs 
in an offline-online framework in an attempt to mitigate the sim-to-real gap, proving the capability of GNNs to generalise well across data distribution and to fit global models. In the context of a robot tasked to re-arrange the shape of a compliant linear object (cable-like), the authors proposed a two-stage learning process. In an offline phase, a GNNs is used to learn the deformation dynamics of the cable purely from simulated data. Then, in the online phase, a linear residual model is learned to minimise the sim-to-real gap between a simulated and the real cable. The trained model is then used to constrain the dynamics of a trust-region -based Model Predictive Controller (MPC) \cite{holkar2010overview}, which computes the optimal robot motion for cable rearrangement. 

Another challenge that may arise when using particle-based models lies on the fact that these often require a temporal consistency between particles. Focusing on the modelling of elasto-plastic objects, in \cite{shi2022robocraft} the authors propose to overcome this limitation by 
training a GNN to model the object dynamics with distance-based supervision between particle distribution, and to predict the deformation of objects when subject to external wrenches (e.g. applied by grippers). 
Interestingly, this representation can be applied in many complex scenarios, such as goal-conditioned rearrangement tasks \cite{deng2022graph, deng2023learning}, where graph-based architectures can be used to directly learn manipulation policies from keypoints, taking as input the keypoints of the current and target state and learning to predict the optimal pick and place set of actions to re-shape the object toward the target configuration. More specifically, two initial sets of detected keypoints are used as nodes of two graphs, encoding initial and target configuration. These are then fed into a local-GNN which implements self-attention operations within each graph, and cross-attention operations across graphs, to perform keypoint matching. These works provide evidence that the use of graphs allows the method to efficiently model the high-dimensional deformable configuration space and its underlying complexity, nonlinearity, and uncertainty, which are intrinsic in deformable object dynamics.

Particles and keypoints are often inferred from visual representation, such as point clouds or mesh \cite{shi2022robocraft}, images \cite{deng2022graph, deng2023learning} and depth images \cite{ma2022learning}. Recently, directly using mesh representations has gained attention as an alternative approach to model complex systems, especially for deformable objects \cite{pfaff2020learning, lin2022learning, mo2023learning, huang2023defgraspnets}. 
Among the others, it is worth discussing firstly MeshGraphNets \cite{pfaff2020learning}, a general framework able to learn the dynamics of a wide range of systems, from cloths to fluids. In a nutshell, the authors propose to encode the state of a mesh, obtained in simulation, in a graph structure that includes mesh-space nodes (i.e. nodes associated with the object) and world-space nodes (i.e. nodes with extra edges representing interaction with external objects and environment). The mesh is processed through an Encode-Process-Decode architecture where the encoder is responsible to build the graph mentioned above, the processor performs message passing along mesh edges and world edges to update nodes embeddings, and the decoder finally computes nodes acceleration (which iterated will produce motion). Interestingly, the authors argue how message passing in mesh-space makes the model learn the internal dynamics of the physical system, while message passing in world-space captures external dynamics such as contacts.
Thanks to this graph-based encoding, MeshGraphNets demonstrated notable capabilities in learning resolution-independent dynamics, enabling variable resolution at runtime, and even an adaptive change of discretization during rollouts, opening to the interesting possibility of allocating larger computational resources to specific local regions where higher accuracy is needed (e.g. corners or contact points). 

In a similar fashion, but in the context of manipulation planning, DefGraspNets \cite{huang2023defgraspnets} propose a multigraph to encode the object mesh together with the gripper mesh, which is then used to learn to predict the deformation of soft objects during manipulation. 

One of the main challenges that arise when learning robotics tasks that imply interaction with deformable objects, such as cloth manipulation, is the intrinsic difficulty in retrieving the object connectivity, because of the large number of degrees of freedom and the self-occlusion conditions. Graphs can be used to tackle this problem and extract directly from visual data the connectivity of a cloth \cite{lin2022learning, mo2023learning}. These methods usually move from a graph built upon the voxelization of an object point clouds, where each node of the graph is the centroid of the corresponding voxel cell, and nodes that are geometrically close to each other (but not necessarily physically) are connected together. A GNN can be used to estimate whether each edge of the graph is also a mesh edge, meaning that it represents also a physical connection of the cloth structure. Then, this visual connectivity graph can be used to learn the model dynamics by means of a GNN that estimates the acceleration of each particle (i.e. of each node). 

\begin{figure}[t]
  \centering  
  \includegraphics[width=0.9\columnwidth]{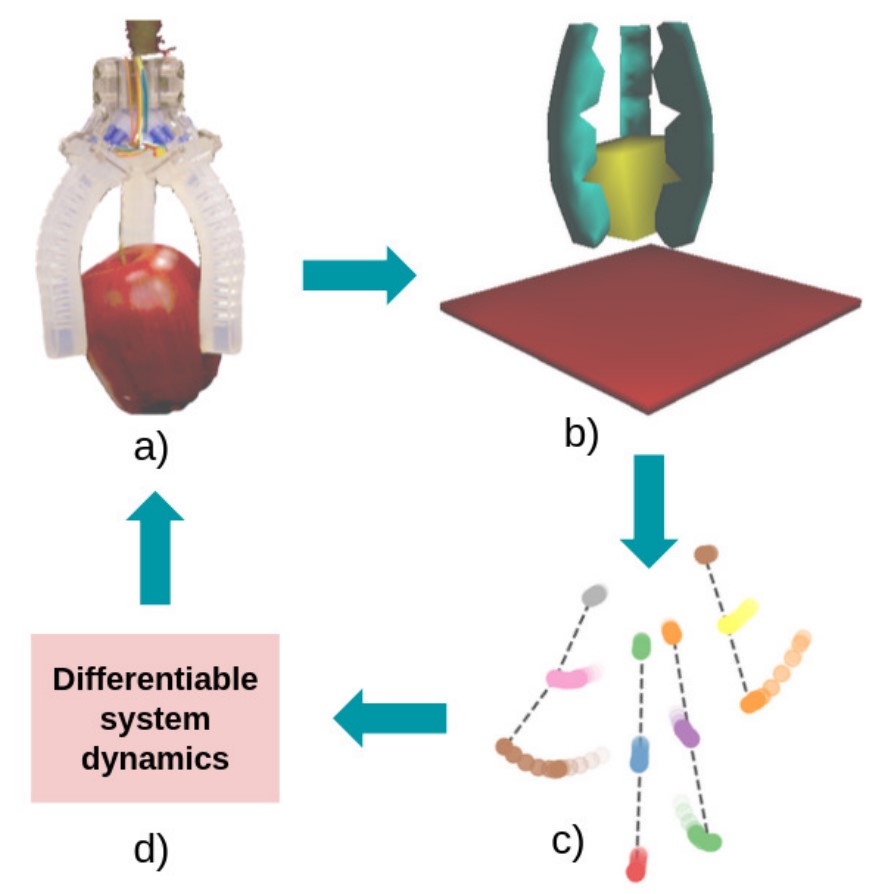}
    \caption{An example of graph used to model a soft end-effector. The GNN is used to learn the dynamics of the system by tracking a selection of keypoints on the robot body. Image taken from \cite{almeida2021sensorimotor}. Copyright \textcopyright 2021, IEEE}
    \label{fig:sensorimotors}
\end{figure}

Graphs can be used not only to model deformable objects but also the robot itself. In general, human, humanoid or animal bodies have a discrete graph structure, where nodes are joints and edges their physical dependencies \cite{wang2018nervenet, oliva2022graph}. NerveNet \cite{wang2018nervenet} builds a graph to represent the structure of the agent, where nodes are different parts of the robot and edges are weighted proportionally to the distance between nodes, and learns continuous agent's control policy by means of a graph neural network, using the implementation of \cite{scarselli2008graph}. The state of each node is updated based on both the aggregated message from its neighbourhood and its current state vector. Such structure has several advantages, the resulting learned policy is general and zero-shot transferable to new scenarios.

In \cite{oliva2022graph}, instead, the authors use a similar model where visual observations are inserted as additional inputs, leading to a graph that encodes both agent and environment status. Both these works prove the importance of training a policy that is aware of the kinematics of the robot. Particularly challenging is the modelling of soft robots, due to their complex continuous dynamics \cite{della2023model}. Related to this topic, it is relevant to mention \cite{almeida2021sensorimotor} (see Fig. \ref{fig:sensorimotors}) where the authors demonstrate how building a graph upon keypoints to model the agent, and using a graph neural architecture, may enable the model to learn to exploit the underlying physical structure.


%


\subsection{Grasping and Manipulation}
One of the primary tasks that an intelligent robot should be able to perform is to safely and effectively interact with objects in its workspace. Indeed, many practical applications require to move, manipulate and use a large variety of objects and tools with different sizes and shapes.

The capability to firmly grasp an object, or to apply controlled wrenches on it with the purpose of changing it pose (manipulation) is one of the primary skills we envision in home-integrated robotic devices. However, although at a first glance this may seem as an easy problem, matching human manipulation skills with modern manipulators is incredibly challenging and a multifaceted problem \cite{prattichizzo2016grasping}. 
%

The development of fully autonomous manipulation systems requires targeting at least three main issues: perception, planning, and control. 
The first refers to the problem of making the robot able to perceive and understand the surrounding scene, identify where (and which) objects are available in the workspace, and how the environment is composed (e.g. static and dynamic obstacles). At this stage, it is particularly important to endow the robot with the capability to learn a representation of the environment and of the objects, which can be used to inform the robot actions for task execution.

Then, once the robot is aware of the scene and of the objects that will be asked to grasp or manipulate, it will be relevant to plan an accurate end-effector motion, which requires the availability of models, usually probabilistic, that predict object state changes when subject to robot actions. 

Lastly, the planned trajectory will be executed with a dedicated controller, which needs to be sufficiently robust to account for disturbances in sensing and for errors in planning and perception, yet at the same time adequately compliant to avoid harsh interactions and damages. 

Albeit being a research problem for decades \cite{bicchi2000robotic, kleeberger2020survey}, in robotic grasping and manipulation several problems are still open, such as learning to interact with objects in motion or in clutter \cite{wen2022catgrasp, berscheid2019robot, dogar2012physics}, manipulation of deformable objects \cite{berenson2013manipulation, yin2021modeling, zaidi2017model, corona2018active,jangir2020dynamic} - for which a proper modelling is crucial as discussed in Sec. \ref{sec:bodies} - and human-robot or multi-robot co-manipulation \cite{li2016human,peternel2016adaptation,dimeas2015reinforcement,elwin2022human}. In addition, when it comes to the deployment of these systems in real-world settings, we may also encounter additional issues: occluded objects due to cluttered scenes or partial observability, novel objects, characterised by new shapes or dimensions, or more sophisticated grasping strategies, for example task-conditioned, where the goal is to select the grasp pose and location depending on the task for which the grasp is performed (i.e. grasping a knife will be executed differently depending whether the robot will cut something, or just handout the tool to someone). 


In the last few years, learning for graphs has provided interesting results in tackling many open problems in this field. Graphs can efficiently model the environment and the interaction between objects  \cite{xie2020deep, lin2022efficient}, learn semantic global information to build knowledge graphs \cite{murali2020taskgrasp, kwak2022semantic, collodi2022learning} or process unstructured data input \cite{neumann2013graph, alliegro2022end, iriondo2021affordance}.
\begin{figure}[t]
    \centering
    \includegraphics[width=0.8\columnwidth]{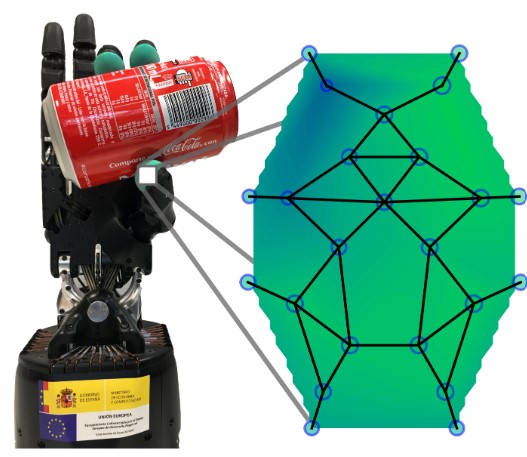}
    \caption{Shadow Robot Dexterous robotic hand endowed with a BioTac SP tactile sensors. Each sensor is represented as graph encoding the continuous sensing surface. Image from \cite{garcia2019tactilegcn}. Copyright \textcopyright 2019, IEEE}
    \label{fig:tactlinegcnn}
\end{figure}

Graph-based structures have been also profitably used to encode spatial relations in complex scenes, where multiple objects are present and/or the environment is only partially observable \cite{zuo2023graph, lou2022learning, lin2022efficient, huang2022planning, ding2022visual, li2020towards, kapelyukh2022my, wilson2020learning, silver2021planning}. 
Object-oriented scenes can be encoded in a graph where nodes are objects and the edges reflect the relations between them, providing a powerful and comprehensive representation of the environment. Notably, since graphs are invariant to the number of nodes, such representation is invariant with respect to the number of objects and therefore can be applied to generic scenes with an arbitrary number of objects.

In real-world scenarios, it is quite common to have cluttered scenes where it is required to perform grasping minimising the probability to cause collisions. The majority of methods usually solve this problem in two steps, first predicting the grasping pose of an object, and then using a collision checking module to verify the safety and feasibility of the planned graph. \cite{mousavian20196, liang2021learning, murali20206, lou2021collision}. In many cases, however, objects are processed independently from the surrounding scene, and the spatial relationship between items is usually disregarded when predicting the grasp pose, which may lead to sub-optimal results and errors in the presence of partial observation.
Yet, several papers demonstrated how building a graph of the objects in the scene enables the exploitation of objects' surroundings knowledge, useful to directly predict effective and safe grasps \cite{lou2022learning, huang2022planning, ding2022visual}. Graph-based encoding showed also interesting features in handling multiple instances of the same item. Assuming to have several similar (or identical) samples of the same object - e.g. in tool boxes, shop carts, or groceries shelves - GNNs are able to capture and highlight the hidden correlation between nodes, which may be exploited for instance to  
select the most accessible sample of the desired object \cite{lou2022learning}. In a similar fashion, \cite{zuo2023graph} 
tackled the problem of grasping partially visible objects. A graph conv in this case is used to uncover and explore non-observable parts of the scene, and to aggregate information extracted from different regions, to capture internal spatial relations and decide whether the target object is already accessible or requires a non-prehensile manipulation primitive (e.g. push) to rearrange its pose.

A relevant topic worth discussing is also the analysis of dynamic multi-object interactions.
\cite{huang2022planning, li2020towards, lin2022efficient} examine long-horizon planning task, where several consequent actions are planned to achieve the desired goal. \cite{huang2022planning} proposes a graph neural network to learn manipulation effects on multiple objects, and proves that the relational inductive bias of this type of network is effective in planning even on a very long time span. Similarly, \cite{ding2022visual} uses the graph relations between objects to predict the grasping order of multiple objects.  In \cite{lin2022efficient} long-term manipulation tasks are seen as a sequence of spatial constraints and objects relationship. The authors proposed to represent the environment state as a graph, where each node encodes features associated with objects and their goals (i.e. target positions). A graph-based architecture is then trained to select the next object to move, where to place it, and predict a task-specific action. 
Interestingly, they also propose an expert demonstration cast as a graph, named GNNExplainer. Their work suggests that graphs are good to encode the knowledge hidden in the examples provided in supervision and not encoded in the problem. This powerful representation of the environment, able to capture spatial dependencies, can be exploited in other more uncommon tasks, as done by \cite{kapelyukh2022my} where the goal was to learn how to place objects in a scene following user’s preferences. Similarly, graphs can also be used to model the interaction between objects and the robot in the context of bi-manual robotic manipulation \cite{xie2020deep}. In this work, the authors introduce a two-level framework composed of a decomposition of the multi-modal dynamics into primitives, and a primitive dynamics model where graph recurrent neural networks are used to capture interactions between objects and the robot arms. In particular, with the graph they model the interaction between different parts of the robot and the manipulated objects. In this case, graph representation helps to improve the performance of the model in several simulated bi-manual robotic manipulation tasks, showing interesting capabilities in increasing training speed and overall accuracy. 

So far we mostly discussed a selection of methods where graphs demonstrated to be convenient tools for modelling environment-level spatial relations between objects and robots.
Interestingly, the same spatial representation problem can be casted to a smaller scale, and model the relationship between fingers and fingertips contacts during grasping (see e.g. Fig. \ref{fig:tactlinegcnn}). Many works, indeed, propose to leverage tactile sensing for the assessment of grasp stability or to foster effective in-hand manipulation \cite{liu2020recent, yamaguchi2019recent,xia2022review, lepora2021soft,xu2022tandem}. However, 
%
although in most cases these sensing strategies leverage cameras to measure sensing surface deformations, their output cannot be easily processed by standard convolutional architectures, because of the complex non-linear relationship between different sensing units. Notably, these issues are discussed and addressed for example in \cite{garcia2019tactilegcn} for tactile sensor state estimation (single sensing unit) and in \cite{funabashi2022multi} for multi-finger pose estimation (multiple sensing unit). The construction of the graph from tactile sensors of \cite{garcia2019tactilegcn} is reported in Fig. \ref{fig:tactlinegcnn} as an example.
%

An insightful application of graphs in the context of robotic grasping and manipulation is the creation of a knowledge graph \cite{tenorth2017representations, saxena2014robobrain, daruna2019robocse, murali2020taskgrasp, kwak2022semantic, daruna2021towards, collodi2022learning, ardon2019learning}, where semantic knowledge and its internal relations are organised in an efficient and powerful way
.

\begin{figure*}[t]
    \centering
    \includegraphics[width=\textwidth]{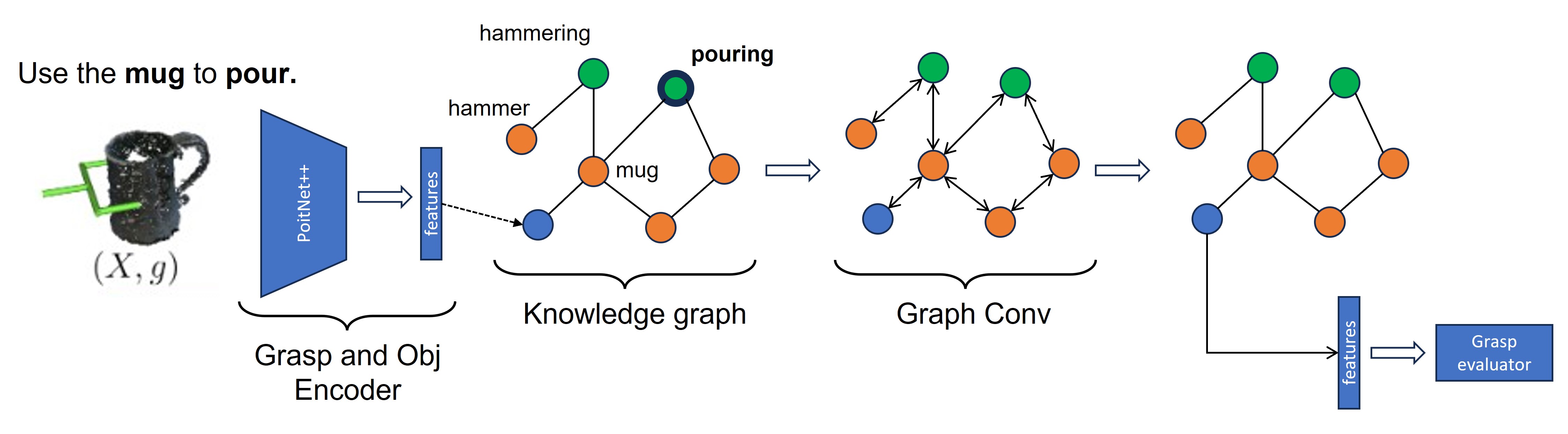}
    \caption{Overview of the method proposed by \cite{murali2020taskgrasp} for task-oriented grasping. It is interesting to observe the utilisation of the knowledge graph to infer information about the query grasp pose. Image redrawn from \cite{murali2020taskgrasp}.}
    \label{fig:TaskGrasp}
\end{figure*}

This structure is able to capture the semantic knowledge about objects, tasks, agents, actions, and regions present in the training environments, and potentially to generalise to novel scenarios, types of objects, shapes, and tasks. Large-scale knowledge graphs such as RoboBrain \cite{saxena2014robobrain} and RoboCSE \cite{daruna2019robocse} propose general frameworks able to extract abstract semantics concepts that can be generalised and take into account the uncertainty of real-world scenarios. Such structure can be exploited for sophisticated robot manipulation, such as for task-oriented grasping first proposed in \cite{murali2020taskgrasp}, where the authors present a framework, named GCNGrasp, where a knowledge graph that encodes objects, tasks and object hierarchies from WordNet \cite{miller1995wordnet} is exploited to decide whether a test grasp pose is suitable for the desired task on that specific object (see Fig. \ref{fig:TaskGrasp}). The classification is done by adding the query grasp pose node on the knowledge graph and by performing few graph convolution layers. Instead, \cite{kwak2022semantic} builds a graph, called RoboKG, with all the available information of the objects, such as material and components, of the tasks, and of the manipulation characteristics, such as type of gripper and exerted force. RoboKG is then used to infer which gripper to use, which part of the object to grasp, and the amount of force required.

Graphs can also be used to encode the functional relationships between grasping strategies, providing a way to contaminate classes representation and to generalise to novel grasping skills. In \cite{collodi2022learning} the authors propose to generate human-inspired grasping strategies depending on objects shapes, claiming that the object shape provide a good affordance for the grasping type, e.g. a cup would be more likely picked with a cylindrical grasping, while a pencil would require a precision pinch grasp. One of the main problems of this representation is the unbalanced dataset, as some strategies are much more rare than others. To overcome this issue, a knowledge graph, built with the knowledge of a kinematic relationship between classes, is used to transfer information between nodes and improve prototypes of long-tail classes.

In robot manipulation, and especially grasping, it is more and more popular to exploit 3D input data, since the depth information is important to a proper grasp prediction
. However, handling 3D data representation such as point clouds is significantly harder than common RGB images, since they lie on non-Euclidean domains, and classic neural network architectures - such CNNs or MLPs - are not effective or even applicable. Graph neural networks have emerged as efficient methods to deal with unstructured data, as demonstrated by a wide literature proposing graph-based architectures to process point clouds \cite{wang2019dynamic, simonovsky2017dynamic,alliegro2022end}. Graph representation of 3D data has already been used in robotics with a straightforward implementation of standard GNNs. In \cite{neumann2013graph} the authors propose a method to improve task-dependent grasping by finding the object category from a given ontology. The authors pose an interesting assumption on the fact that objects with similar shapes would be grasped in a similar way for similar tasks. Thanks to the representation of each 3D object as a graph, where nodes are points and edges are weighted by the changes of the surface, graph kernels are used to compare the similarity between objects, with the purpose of fostering knowledge sharing between objects of similar shape.

\subsection{Human Action recognition}
Human action recognition, tracking, and forecasting are rapidly becoming a key enabling factor for a variety of robotics applications, ranging from technology-enabled healthcare, all the way to assistive robotics and human-machine interaction and cooperation. As an example, it is worth mentioning gesture recognition, where the robot is tasked to detect and interpret human hand and arm movements, to favour a natural interaction \cite{kong2022human}. This can be used in a wide variety of different ways, for example, a robot in a healthcare facility could be able to recognise specific hand gestures made by patients or doctors, and therefore trigger specific reactions, e.g. fetch tools or program actions  \cite{wang2019survey}. Human gesture understanding can be also used to feed learning-by-demonstration or imitation learning pipelines \cite{argall2009survey, ekvall2008robot, fang2019survey}, with the goal of transferring human skills to manipulators, making them able for example to fold a towel \cite{tsurumine2022goal}, paint a piece of furniture or manipulate kitchen items \cite{wang2023mimicplay}. 

Beyond making robots more and more autonomous for daily living tasks, an open and - in some cases - more challenging problem is to enable a safe physical interaction with the environment and the humans \cite{albini2017human, de2008atlas}. In this case, human action recognition becomes even more crucial, because the robot may benefit from understanding whether people in its surrounding will walk or move their limbs in specific portions of the workspace, to either avoid contacts or trigger consequent support actions (e.g. pass a tool or co-manipulate an object) \cite{alahi2016social, huang2019stgat, thobbi2011using}. Of note, understanding human behaviour, and eventually being able to react consequently, may have a significant effect on the anthropomorphism and accountability of the machine, which proved to be a key factor for the increased acceptability and trustworthy of machines in our daily life \cite{natarajan2020effects}.

The use of graph neural networks for human motion modelling has a longstanding tradition, especially for body keypoints estimation \cite{wang2020graph, jin2020differentiable, qiu2020dgcn, peng2020learning}. Several surveys have been proposed to summarise the state of the art of deep learning methods for skeleton-based human pose reconstruction (as for example \cite{wang2021deep}), and therefore will be here omitted to preserve the robotics-oriented focus of this paper and to leave more space to the modelling of complex human behaviour when interacting with objects and the surrounding environment. 

Human behaviour is an intricate and multifaceted phenomenon, strongly affected by individual personality traits, past experiences, cultural background, and social influence, which are interleaved in a unique and unpredictable way. These aspects play a role even in controlled scenarios, e.g. during a meal preparation we may choose different steps order depending on our habits. To master this complexity, graphs appear to be a valid choice, thanks to the possibility to represent in a compact fashion the functional or semantic relationship between knowledge entities encoded in nodes. Under the domain of vision-based systems, which represents the primary sensing modality for this kind of application, graphs can be used to model structured information such as zones of an environment (the sink, the oven and the fridge in a kitchen, the desk, the armchair and the window in an office), or contacts between hands and objects \cite{ahmad2021graph, nunez2022egocentric}. This line of research has been particularly fostered by the release of large unstructured datasets, among which it is worth mentioning Kinetics \cite{carreira2017quo} and Charades \cite{sigurdsson2016hollywood} for third-person videos, and Epic Kitchens \cite{damen2018scaling},  Ego-4D \cite{grauman2022ego4d}, and Assembly101 \cite{sener2022assembly101} for egocentric vision. 

\begin{figure}[t]
    \centering
    \includegraphics[width=\columnwidth]{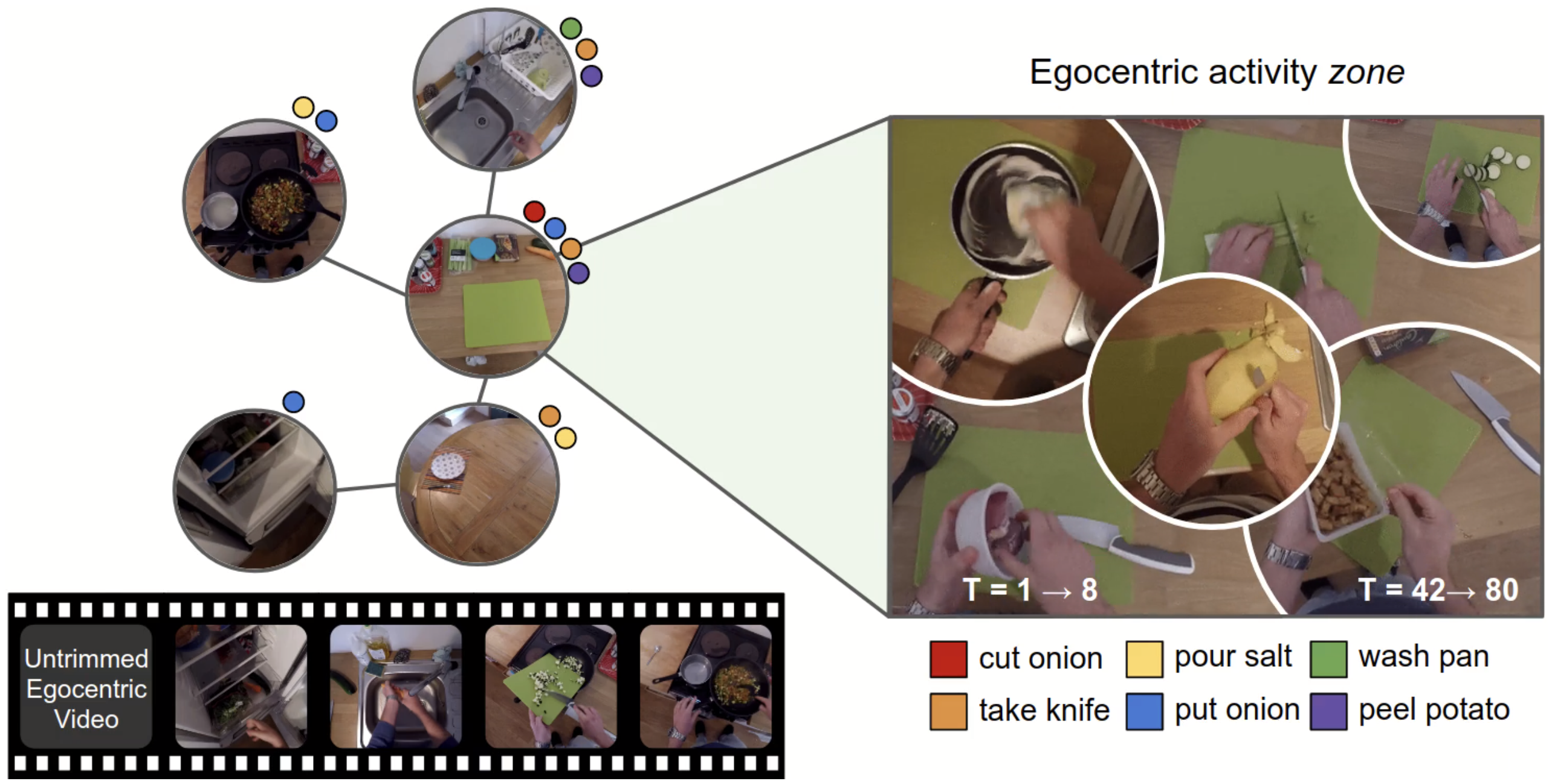}
    \caption{Topological graph of activity-centric zones. Image from \cite{nagarajan2020ego}. Copyright \textcopyright 2020, IEEE.}
    \label{fig:egotopo}
\end{figure}

One notable example is \cite{nagarajan2020ego}, where the authors discuss the feasibility to detect relevant zones in a domestic environment from egocentric videos (activity-centric zones) and encode them in a graph structure where new zones explored are sequentially added to the graph. Interestingly, such topological representation of environmental key zones, reported for reader's convenience in Fig. \ref{fig:egotopo}, intrinsically embeds information on which actions may take place in specific zones (i.e. if the camera is looking at a sink, the user will likely wash something), while edges capture spatial relationships between different zones, depending on how people navigate between them. Topological graphs like the one proposed in \cite{nagarajan2020ego} demonstrate an interesting capability to support deep learning models for action forecasting, where the graph is used to encode a series of zones visited during a complex action (e.g. the user is first at the sink, and then move to the cutting board) to predict the next activity (e.g. cut something). Interestingly, this source of knowledge could be pivotal to enable a more natural and efficient learning of human skills with minimum motor impairment, opening interesting perspectives for imitation learning -based strategies to transfer human action skills to manipulators, or to inform a robot of the next human action, for contact avoidance and cooperation. 

With a similar yet different purpose, graphs represent convenient mathematical structures to model human-object interactions. As an example, in \cite{dessalene2020egocentric,dessalene2021forecasting} the authors propose a graph to encode contacts between hands and objects. The nodes state contains the information of which object is in contact with the two hands in the current and future time-spans, with the purpose of anticipating future actions of the user. A complex activity is then represented by a sequence of nodes, where edges represent temporal relationships (before/after) providing temporal context to the action recognition neural architecture. 

In \cite{sieb2020graph}, the authors tackle the problem of visual imitation, meaning the ability of an agent to learn skills by observing and imitating a human demonstrator. Learning from complex and unstructured human behaviours requires a fine-grained understanding of the demonstrator's visual scene and of how it changes over time, which is addressed in \cite{sieb2020graph} through the definition of hierarchical graph video representations, named Visual Entity Graphs (VEGs). The general idea is to encode human behaviour through a set of nodes representing visual entities - such as objects or hands - tracked in space and time, and connected through edges encoding their relative 3D spatial arrangement. Such structured representation enables a more fine-grained understanding of the activity, the key visual entities in the scene, and their spatial arrangement in the demonstrator's and imitator's environments. This encoding enables the agent to better understand the visual scene and changes over time, which is essential for successful visual imitation. The graph structure also allows for more efficient computation of visual similarity between the demonstrator's and imitator's environments, which is maximised through the imitator's actions to result in a correct transfer of skills.

Of relevance for this topic is also \cite{doosti2020hope}, where the authors address the challenges of estimating from videos the three-dimensional poses of hand and grasped objects in real-time during manipulation tasks. In this case, the graph encodes the spatial relationship between hand joints and object corners, and can dynamically adjust its structure depending on the input data, allowing the model to handle different hand-object configurations. Interestingly, the authors demonstrated that GNNs can make the difference in both refining the estimation of 2D keypoints, and also in converting them into their final 3D representation, also enabling on-line processing of visual streams.

\section{GNNs for Multi-Agent systems}\label{sec:multi}
In multi-agent systems, an effective coordination and communication between robots is of paramount importance  to enable a fruitful cooperation, which is clearly critical for several downstream tasks. Traditionally, this problem has been addressed by exploiting centralised approaches, where all the intelligence and computation complexity is centralised on a common controller node, which collects information from the environment, plan and then communicate single actions to each robot to fulfil the desired task. Such an approach has the strong benefit of having a central control system able to reason on the current (and past) status of the whole set of agents and the surrounding environment, thus requiring only little computation on the device side for action execution. However, this approach has a low fault tolerance: if even a single agent fails, all the actions of the robots need to be recomputed. Centralised solutions also badly scale with the increase of agents, and for difficult tasks it even requires additional per-robot computation \cite{Khan2020GraphNN}. 

To address these issues, recently many researchers devoted their attention to decentralised approaches, where each agent learns its set of actions according to environmental data and shared information from other agents. In this configuration, each robot has to deal with limited observability due to the range of the sensors, each robot can only see a partial portion of the environment and can communicate with the nearest agents. 
GNNs offer generalized and flexible structural representations of the elements in multi-robot systems, and have been recently widely used in decentralised approaches to adress several tasks. 

The utilisation of GNNs can change according to the information that are encoded in the graph. 
Many notable works \cite{sun2020scaling, tolstaya2020learning, wang2020learning, pmlr-v155-blumenkamp21a, kortvelesy2021modgnn, clark2021queue, wang2022heterogeneous, sebastian2022lemurs, gama2022synthesizing, gosrich2022coverage, tzes2022graph} focus the attention on the communication between robots, a crucial aspect for task and motion planning, and exploit graph encoding to efficiently model inter-agents relationships. Usually, graph neural networks are used to learn a communication graph, which is then exploited for example to directly support the agents in making decisions, or to foster multi-agent reinforcement learning frameworks in policy learning. 

On the other hand, several methods propose to encode the information related to the environment topology, in the form of waypoints or point-of-interest \cite{chen2019autonomous, chen2020autonomous,chen2021zero, agarwal2020learning, zhang2022h2gnn, tolstaya2021multi}. Works as \cite{chen2019autonomous, chen2020autonomous, chen2021zero} focus on the spatio-temporal relationship between entities in the environment to build a graph, and use GNNs to learn environmental features for the robot. Other methods \cite{agarwal2020learning, zhang2022h2gnn}, instead, propose a more comprehensive topological representation where both robots and environments are directly encoded in the graph. In particular, \cite{agarwal2020learning} is the first work that introduces a shared agent-entity graph in the context of multi-agent learning. Both environments entities and agents are encoded as nodes in the graph, and the edges reflect the communication links, which can be assumed to be always available (fully connected graph) or constrained by some sort of maximum distance range (i.e. edge exists only if the distance between nodes is below a given threshold). The communication inside the graph is performed in two steps: first, each agent computes its state embedding and aggregates the environment information with an attention mechanism \cite{vaswani2017attention}, then, an inter-agent communication is performed to share local information along the team members.

Generally speaking, graphs provide several desirable advantages for multi-robot systems: 
i) invariance to the number of agents/points of interest ii) invariance to the permutation of the agents iii) suitability to learn policies for Multi Agent Reinforcement Learning (MARL) and imitation learning iv) policies learned on small systems are transferable to systems with a larger amount of agents. In the following, we will showcase how these benefits have been exploited for two relevant multi-robot scenarios: task and motion planning, where a fleet of robots move coordinately in an environment and cooperates to accomplish one or more activities; exploration and navigation, where agents are tasked to explore an unknown environment.


\subsection{Task and Motion Planning}
Graph structure is a natural way to describe decentralised logic. This implementation provides a way to plan multi-vehicle trajectories with performances similar to centralised approaches but in a decentralised manner, with a massive reduction of problem complexity and communication burden.
In multi-robot path planning, the goal is to find collision-free paths to bring each agent in the team from a starting position to the target one. 

The task can be particularly challenging when the agents have constraints in terms of range of observability and communication, quite common in real-world applications. Communication between agents is always crucial in multi-robot systems, and to fulfil this task it is particularly important to decide which information to share, when, and how, so that each agent is able to take effective decisions. To address this problem, \cite{li2020graph} proposes to mix a convolutional neural network and a graph convolutional neural network to extract visual features from local observations of each agent, and to exchange such information within the team respectively. The graph neural network is structured as in \cite{prorok2018graph}, where each agent is a node and the edges are communication connections determined by a distance threshold. Such joint configuration helps the network to identify the relevant information to share with the team, in order to efficiently perform path planning. The work is later extended in \cite{li2021message}, and an attention-like mechanism \cite{velivckovic2017graph} is inserted in the graph neural network to model the intra-robot communication, where the weights on edges between nodes are proportional to the importance of the received message. 
A similar structure is exploited by \cite{li2023dynamic}, where an history of past paths is stored within the graph neural network to perform motion planning. 

\begin{figure*}[t]
    \centering
    \includegraphics[width=\textwidth]{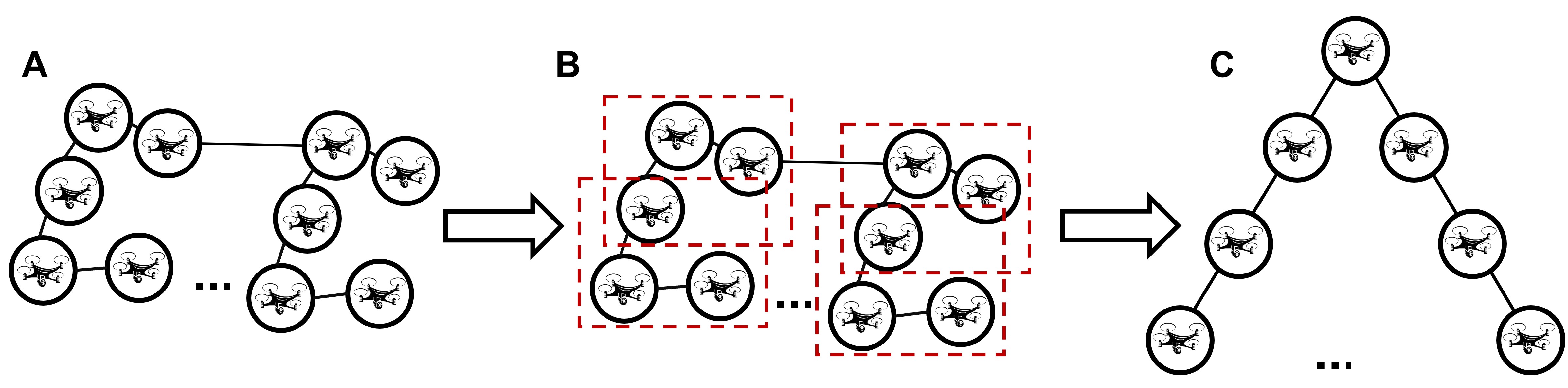}
    \caption{Example of a communication graph used to learn policies to produce a desired behaviour. Redrawn from \cite{khan2020graph}.}
    \label{fig:comm_graph}
\end{figure*}

One challenge trend of research in multi-robot systems is the scalability to large numbers of agents, mostly because of the increasing communication burden. Graph neural networks offer a good solution to this issue, because they can mitigate the computation cost by aggregating information locally and not throughout the entire system. Interestingly, a few works also demonstrated the feasibility to learn over a small number of agents a general model, which can be seamlessly extended to larger fleets \cite{khan2020graph, khan2019graph, khan2021large}. Such works introduce Graph Policy Gradients, an algorithm that exploits graph neural networks to learn policies for the robots. A GNN is trained in the classic configuration for multi-robot systems, where each node is an agent and the edges connect team members that are within a threshold distance. Such a network can then be applied to several problems such as formation flying and path planning (see e.g. Fig. \ref{fig:comm_graph}). The paper also discusses the scalability of GNNs in this framework toward an arbitrary number of robots, showing interesting perspectives in terms of limitation of computational cost and ability to transfer the learned policy to larger fleers in zero-shot, remarkably training on systems with cardinality from as little as three robots all the way up to a hundred.

A task that shares some similarities with path planning is to learn scheduling policies, which consists of coordinate agents programmed to complete tasks in predefined times and locations. To overcome similar issues of path planning, namely the difficulty to scale to large systems and lack of generality, GNNs have been investigated as a way to address them in RoboGNN \cite{wang2020learning}, where the authors propose a graph attention -based model able to solve simple temporal network-based scheduling problems. 


\subsection{Exploration and Navigation}

In multi-robot exploration, a team of robots needs to explore an unknown environment and additionally visit several regions of interest. In such a scenario, it is particularly important the communication and cooperation between agents to efficiently cover the exploration ground, and avoid conflicts or repeated explorations. Furthermore, in order to successfully fulfil the task, a key aspect is to define the spatial relationship between robots and regions of interest and extract environmental information. Graph neural networks provide a promising tool to efficiently regulate the communication between agents and aggregate topological information from the environment. 

In section \ref{sec:multi}, we reviewed several approaches that build a communication graph for multi-robot systems coordination, which can be also applied for exploration tasks \cite{sun2020scaling, tolstaya2020learning, wang2020learning, pmlr-v155-blumenkamp21a, kortvelesy2021modgnn, clark2021queue, wang2022heterogeneous, sebastian2022lemurs, gama2022synthesizing, gosrich2022coverage}. Some of these works, instead, are specifically designed for the task of exploration, such as \cite{gosrich2022coverage} where the authors tackle the task of coverage control, with the aim of predicting the distribution of a set of robots in a region such that the likelihood to spot events of interest is maximised. Of relevance is also \cite{clark2021queue}, where the communication graph between several agents and a data sink is used to prevent data loss and allow robots to explore the environment efficiently in the presence of intermittent connectivity, for instance in space exploration.

It is also relevant to mention a line of research where graphs are used to model together both the environment and the agents therein. 
For the sake of completeness, we first introduce some relevant works which target this problem for single-robot exploration.
In \cite{chen2019autonomous, chen2020autonomous, chen2021zero} the authors propose an exploration graph to predict the next action \cite{chen2019autonomous}, further extended with the addition of deep reinforcement learning to learn robot policies \cite{chen2020autonomous, chen2021zero}. The model provides a general representation of the SLAM-dependent robot state and environment, creating a spatio-temporal graph where robot state, landmarks, and frontiers are encoded as nodes, and a graph neural network is used to extract meaningful features from the environment. Topological graphs can be also used to model an entire environment, such as in \cite{gomez2019topological}, where the graph is used to model an indoor space for mobile robot exploration. 
The idea of processing environmental information is then later extended also to multi-robot exploration \cite{luo2019multi, tolstaya2021multi, zhang2022h2gnn}. In \cite{luo2019multi}, the authors propose to segment the exploration environment into the graph domain. To do this, the map is divided into regions and each of them is mapped into a node of the graph, while edges between nodes represent spatial distance. Based on the topological graph, a graph-based reinforcement learning algorithm is applied to learn how to assign exploration targets to each robot. Instead, in \cite{tolstaya2021multi, zhang2022h2gnn} the authors introduce the use of a spatial graph, where both map locations and agents are represented as nodes, and the connectivity encodes the set of allowed moves. In particular, \cite{tolstaya2021multi} uses behaviour cloning to train a GNN controller to imitate the expert solution. Such a model can be easily generalised to scenarios with a larger map and more agents.
The usage of a GNN allows to easily choose the range in which the exchange of information may happen, by just varying the receptive field of the network.

In \cite{zhang2022h2gnn}, Zhang et al. propose a coarse-to-fine exploration framework based on graph neural networks. Depending on the type of exploration, different regions of the environment may have different degree of relevance. As an example, for coarse exploration, the boundary is probably the most relevant zone where to focus, while for a fine inspection, looking to closer regions could be more informative. Graph neural networks can easily reflect such necessity giving more or less importance to features coming from different parts of the graph.


\section{Strengths and Limitations of current literature}
Moving from our analysis of the literature, we believe that it is relevant to mention how graph-based representations opened interesting research perspectives, such as for the modelling of functional, spatial, or temporal relationships between passive or active elements (between robots, objects, and zones of an environment).
Interestingly, in the last few years, we observed an exponential increase in the adoption of this learning method for robotics applications, which at the moment is consolidated as one of the most prominent, together with images, for complex tasks. 

For the field to progress to a higher maturity phase, however, we strongly believe that some aspects still deserve additional work. First, the analysis we performed strongly motivates and foster a more extensive and informed use of graphs as a convenient tool to represent unstructured information and enable learning on complex data structures. Yet, in many cases, graph learning is used in a naive fashion, albeit specific applications may benefit from advanced graph neural architectures formalisation such as attention based (see Fig. \ref{fig:gnns_formalization}). One of the purposes of this survey is to bridge this gap by collecting and providing a cheat sheet on the most advanced graph learning theory for the robotics community.

\section{Conclusions and future perspectives}
The main goal of this paper is to showcase a comprehensive overview on the use of neural networks for graphs in robotics applications. Indeed, in the last decade, roboticists highly benefited from the advancement of techniques and methods defined in the machine learning community, serving as one of the preferred test benches for deep learning. However, often practical robotics use-cases deal with complex and unstructured data, such as three-dimensional data (point clouds), functional and temporal relationships between elements, etc. To unlock the potential of machine and deep learning for these applications, it is crucial to make models able to process unstructured data representation, for which graph encoding seems to be the preferable approach. We revised and categorised more than 100 papers on the topic, trying to identify the major trends of research in robotics in which graph encoding and learning play a crucial role in enabling novel tasks and making efficient the learning process. 

Our literature review suggests that graph learning can enable a plethora of novel tasks, ranging from action recognition and forecasting, to space representation all the way to soft-bodies modelling. Interestingly, the number of novel applications is increasing with time, and we hope that our effort of collecting and revising the major contributions to the topic may provide a boost to the research on robotics-enabled novel and more complex tasks. 

Finally, we also believe that the robotics community may not only serve as a test bench for machine learning, but could also contribute actively to the development of the field. Realistic problems coming from the robotics field could provide insights, constraints, and guidelines to foster novel learning and modelling scheme for researchers working on the foundation of graph learning. For this reason, this survey is also addressed to machine learning scientists, delivering a complete overview of the major challenges that the robotics community is currently addressing, with the hope of providing interesting research hints to bridge the gap between theory and application. 


\section*{Acknowledgment}
This study was carried out within the FAIR - Future Artificial Intelligence Research and received funding from the European Union Next-GenerationEU (PIANO NAZIONALE DI RIPRESA E RESILIENZA (PNRR) – MISSIONE 4 COMPONENTE 2, INVESTIMENTO 1.3 – D.D. 1555 11/10/2022, PE00000013). This manuscript reflects only the authors’ views and opinions, neither the European Union nor the European Commission can be considered responsible for them. 
%

\bibliographystyle{IEEEtran}
\bibliography{biblio}

\begin{IEEEbiography}[{\includegraphics[width=1in,height=1.25in,clip,keepaspectratio]{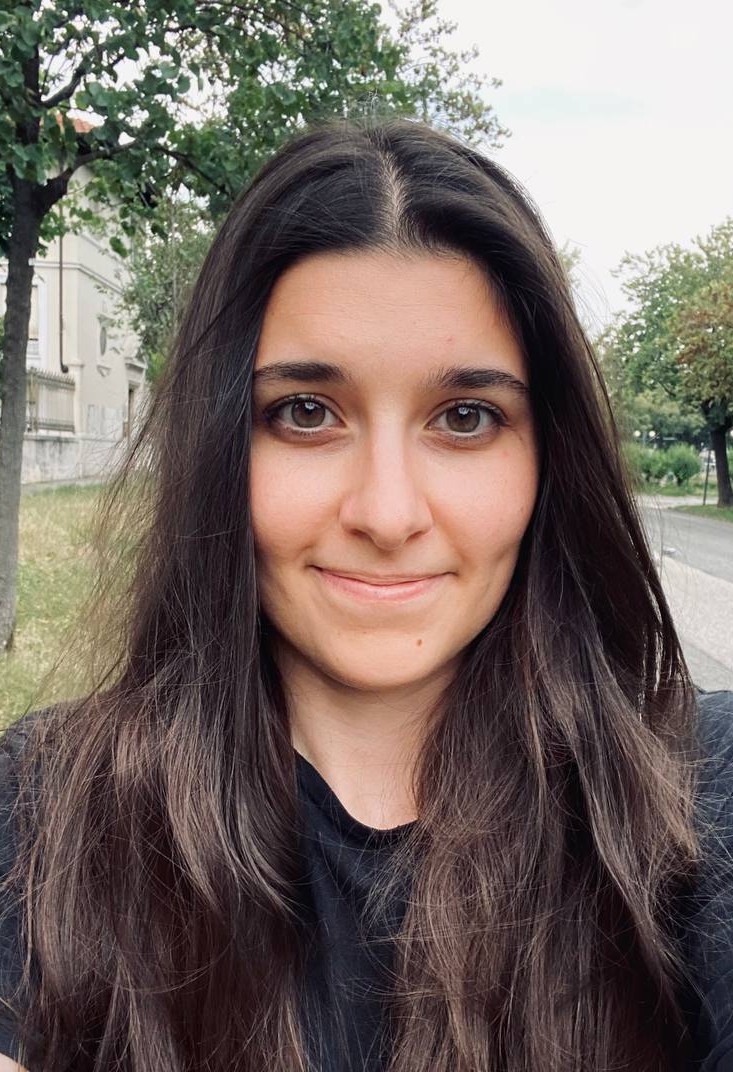}}]{Francesca Pistilli} (Student Member, IEEE) Francesca received the PhD degree at the Polytechnic of Turin at the Image Processing and Learning group (IPL). She previously received the M.Sc. degrees in Electronic Engineering from Polytechnic of Turin and Electrical and Computer Engineering from the University of Illinois at Chicago, Chicago, IL, in 2019 and 2020 respectively. She is currently a postdoctoral researcher in computer vision at the Polytechnic of Turin. Her interests include graph-convolutional neural networks, implicit neural representations, transformers, and other cutting-edge deep learning algorithms applied to image, point cloud processing and robotics.
\end{IEEEbiography}

\begin{IEEEbiography}[{\includegraphics[width=1in,height=1.25in,clip,keepaspectratio]{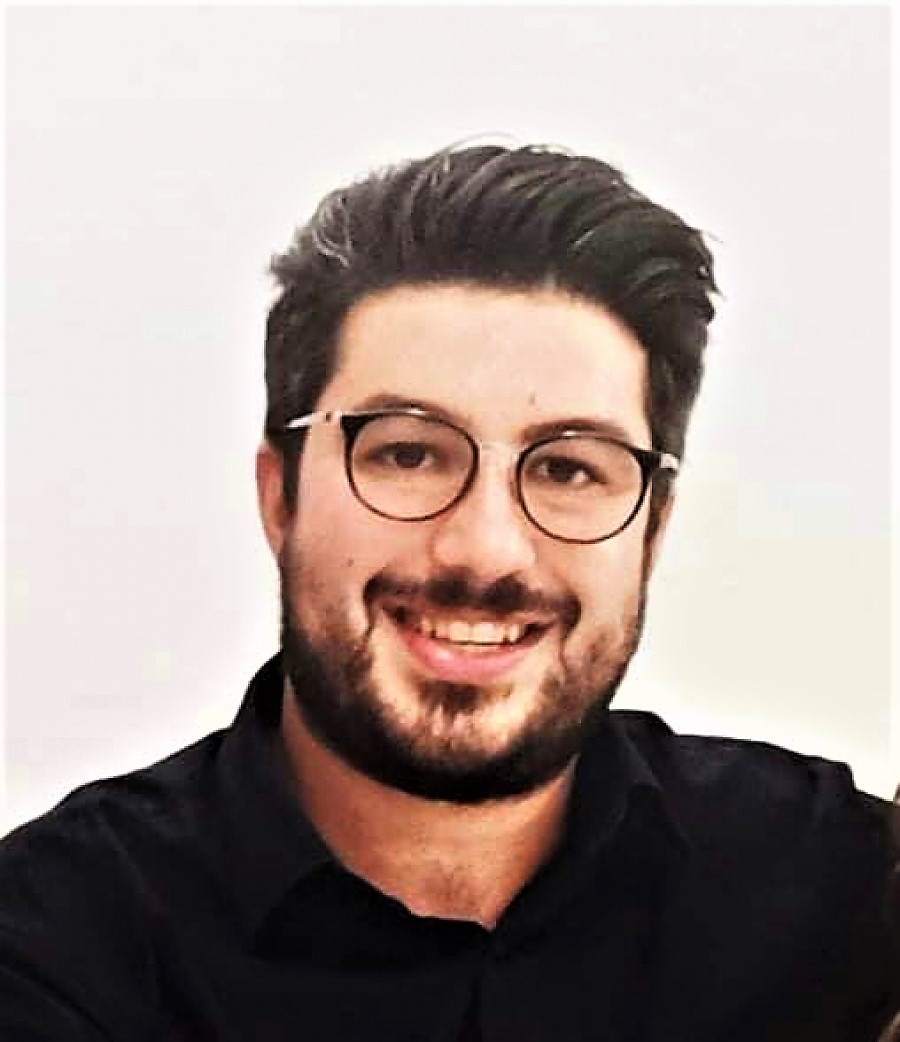}}]{Giuseppe Averta} (Member, IEEE) Giuseppe is currently Assistant Professor of Robotics and Machine Learning at the Polytechnic of Turin. Previously, he was a postdoctoral researcher at the University of Pisa, where he got his B.S. in Biomedical Engineering and his M.S. in Robotics, with honors, in 2013 and 2016, respectively. He is also an Italian Institute of Technology Alumnus. In 2019, he was a visiting student at the Eric P. and Evelyn E. Newman Laboratory for Biomechanics and Human Rehabilitation Group at MIT. Giuseppe’s research interests are in the development of a truly embodied intelligence for human-robot cooperation, with research activities in the optimization of machine learning models for edge computing, deep learning for egocentric vision, and on human-inspired design, planning, and control guidelines for autonomous, collaborative, assistive, and prosthetic robots.

\end{IEEEbiography}

\EOD

\end{document}